\documentclass[iicol]{sn-jnl}

\usepackage{graphicx}%
\usepackage{multirow}%
\usepackage{amsmath,amssymb,amsfonts}%
\usepackage{amsthm}%
\usepackage{mathrsfs}%
\usepackage[title]{appendix}%
\usepackage{xcolor}%
\usepackage{textcomp}%
\usepackage{manyfoot}%
\usepackage{booktabs}%
\usepackage{algorithm}%
\usepackage{algorithmicx}%
\usepackage{algpseudocode}%
\usepackage{listings}%
\usepackage{subcaption}
\usepackage{natbib}

\begin{document}

\title[Article Title]{An HMM-based framework for identity-aware long-term multi-object tracking  from sparse and uncertain identification: use case on long-term tracking in livestock}

\author[1]{\fnm{Anne Marthe Sophie} \sur{Ngo Bibinbe}}\email{anne-marthe-sophie.ngo-bibinbe.1@ulaval.ca}

\author[2]{\fnm{Chiron} \sur{Bang}}\email{fbangnjenjoc2021@fau.edu}

\author[3]{\fnm{Patrick} \sur{Gagnon}}\email{pgagnon@cdpq.ca}

\author[1]{\fnm{Jamie} \sur{Ahloy-Dallaire}}\email{jamie.ahloy-dallaire.1@ulaval.ca}

\author*[1]{\fnm{Eric R.} \sur{Paquet}}\email{eric.paquet@fsaa.ulaval.ca}

\affil[1]{\orgdiv{Animal Science Department}, \orgname{Laval University}, \orgaddress{\city{Quebec City}, \state{QC}, \country{Canada}}}

\affil[2]{\orgdiv{Department of Electrical Engineering and Computer Science}, \orgname{Florida Atlantic University}, \orgaddress{\city{Boca Raton}, \state{FL}, \country{USA}}}

\affil[3]{\orgname{Centre de développement du porc du Québec}, \orgaddress{\city{Lévis}, \state{QC}, \country{Canada}}}


\abstract{The need for long-term multi-object tracking (MOT) is growing due to the demand for analyzing individual behaviors in videos that span several minutes. Unfortunately, due to identity switches between objects, the tracking performance of existing MOT approaches decreases over time, making them difficult to apply for long-term tracking. However, in many real-world applications, such as in the livestock sector, it is possible to obtain sporadic identifications for some of the animals from sources like feeders. To address the challenges of long-term MOT, we propose a new framework that combines both uncertain identities and tracking using a Hidden Markov Model (HMM) formulation. In addition to providing real-world identities to animals, our HMM framework improves the F1 score of ByteTrack, a leading MOT approach even with re-identification, on a 10-minute pig tracking dataset with 21 identifications at the pen's feeding station. We also show that our approach is robust to the uncertainty of identifications, with performance increasing as identities are provided more frequently. The improved performance of our HMM framework was also validated on the MOT17 and MOT20 benchmark datasets using both ByteTrack and FairMOT. The code for this new HMM framework and the new 10-minute pig tracking video dataset are available at:
\href{https://github.com/ngobibibnbe/uncertain-identity-aware-tracking}{https://github.com/ngobibibnbe/uncertain-identity-aware-tracking}
}


\maketitle
\section{Introduction}
\label{sec:intro}

Multi-object tracking (MOT) involves localizing and uniquely identifying objects in a video. While existing approaches succeed in performing MOT for short videos, their performance deteriorates significantly over time on longer videos \citep{Luo2021,valmadre2018long}. This is primarily due to inherent tracking challenges like object identity switches and loss of tracking identities caused by erratic object movements, occlusions, and the presence of objects with similar appearances \citep{Luo2021,valmadre2018long}. Existing MOT approaches are usually supervised \citep{10030267,zhang2021fairmot}, self-supervised \citep{Bastani2021, Vondrick2018}, or unsupervised \citep{Zhang2021,bewley2016simple,Wojke2018}. Among these, there is ByteTrack \citep{Zhang2021} that performs tracking by object detection with the particularity of keeping objects of low detection score in a buffer (eg. occluded objects) to match them, if necessary, with similar tracklets. There is also FairMOT \citep{zhang2021fairmot} which learns at the same time the object detection, tracking, and Re-ID features through the detection and Re-ID heads of its network.  Most approaches have been tested on benchmarks such as \citep{8953401} and \citep{peize2021dance} which contain videos lasting usually less than 3 minutes. 
In many applications of MOT, tracking is often required over long periods, ranging from several minutes to hours. For example, in animal behavior analysis, livestock management, video surveillance, and sports analysis, MOT is often needed for hours-long durations \citep{psota2022,cui2023sportsmot}, making 3-minute videos relatively short as a benchmark.

Furthermore, in those MOT applications, it is generally essential not only to detect and track objects but also to assign them to their real-world identities (RWID, e.g., animal IDs, person names, etc.)
and to track those RWID long enough for further analysis (e.g. detect health problems or aggressive behaviors) \citep{psota2022, marsot2020adaptive}. Assigning real-world identities to objects is known as identity-aware MOT, and RWID could be provided by external sources like facial recognition algorithms \citep{7298718} or automated identification stations as it is the case in the livestock sector \citep{psota2022,Yu2016} with feeders and drinkers. Livestock animals could also be equipped with special markers \citep{psota2022} (e.g. tags, QR codes) that are easy to assign to RWID directly from videos, but those markers are rarely used and require extra maintenance. The ideal situation in this sector is to be able to obtain RWID from readily available tools like animal feeders/drinkers where animals are identified through already attached radio frequency identification (RFID) tags used to monitor their feeding/drinking behavior.
\begin{figure*}[h!]
\begin{center}

\includegraphics[width=16cm]{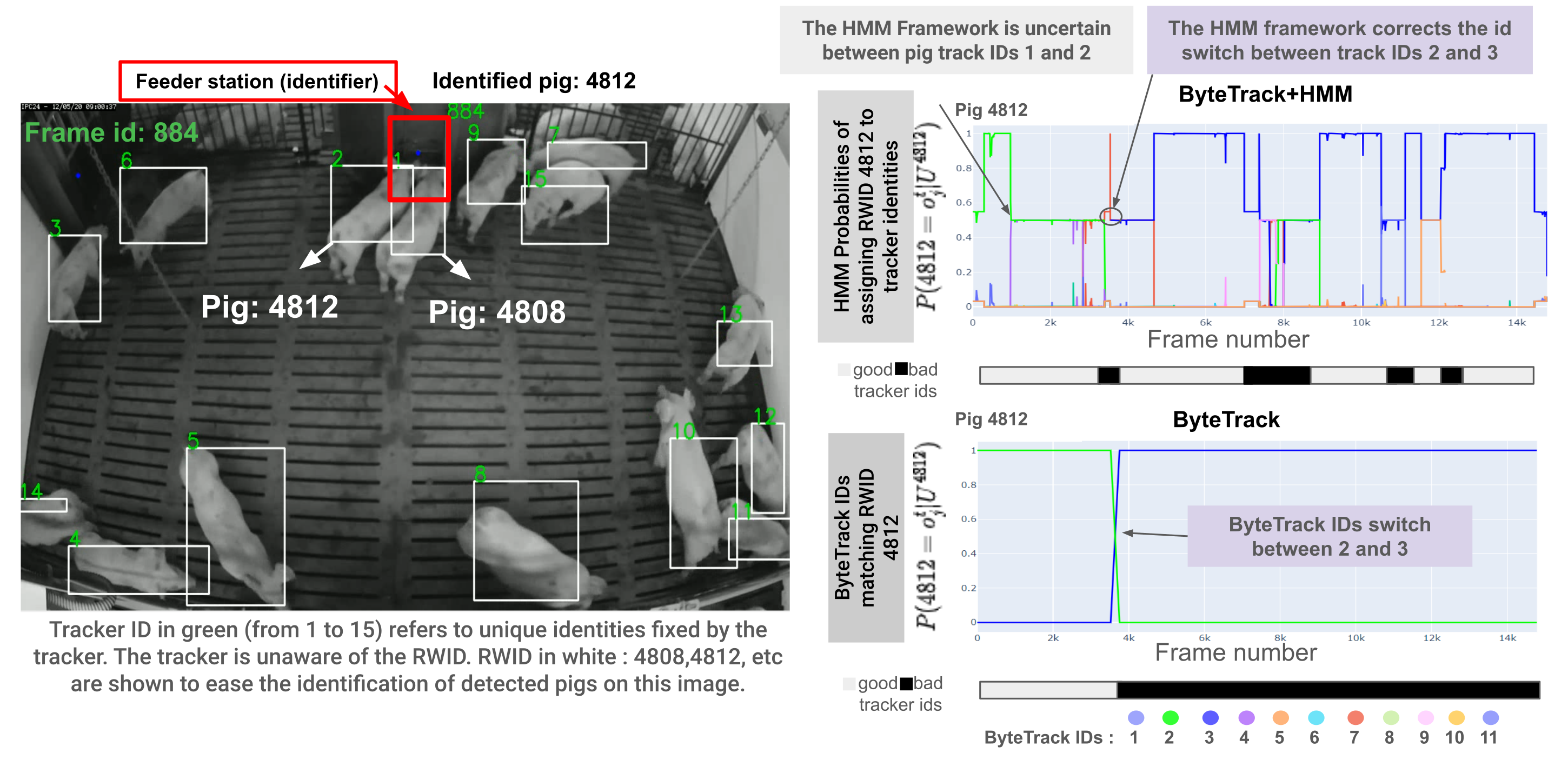}
\caption{(Left) On frame 884 of this video, the pig with the RWID 4812 is detected at the feeding station (the red rectangle at the top). At the same time, the tracker ByteTrack detected two pigs with IDs 1 and 2 near the feeding station. In this example, the feeding station is the $identifier$ (i.e. the device providing the official RWID)}. Each pig gets a 50\% chance of being RWID 4812. The HMM framework then splits the chance between them (upper right). Later at frame 3666 of the same video, ByteTrack switched IDs 2 and 3 (lower right). Our approach ByteTrack+HMM, using the $identifier$ past and future information, corrects this by assigning 4812 to ID 3 after frame 3666 instead of ID 2.
  
   \label{fig:station}

   \end{center}
\end{figure*}

\section{Related Works}
In order to address long-term identity-aware tracking, information on identities is used for re-identification \citep{Yu2016,7298718}. Re-identification consists of correcting the tracking by using identifier (i.e. entity providing RWID identification information) inputs where the identity provided by the identifier is assigned to the object. Here the identifiers could be facial recognition models, active learners, feeding/drinking stations, etc. However, those approaches rely on the accuracy of the identifier to identify the detected objects. Those approaches do not take into account contexts where identifications are sparse over time. \citep{7780789} formulated a solution for identity-aware tracking merging tracking by detection and RWID information from an identifier. They formulated the tracking problem as a quadratic optimization problem with specific constraints ensuring that a detected object can only be associated to one RWID and vice versa. Unfortunately, for this framework, the identities are certain since they are based on face recognition. This framework is thus not suitable for a context where identities could be assigned to more than one nearby objects due to uncertainty.

While those methods could be suitable for identity-aware MOT, they do not necessarily take into account the uncertainty of the identifier, and the proposed solutions are not exact solutions but approximations. However, uncertain identifications are frequent in various applications, such as in the livestock sector where animals are identified at the feeding stations using RFID tags \citep{voulodimos2010complete}. For example, there could be many animals near the feeding station, as presented in Figure \ref{fig:station}. This frequent situation leads to uncertain identifications of nearby animals.
\citep{Camilleri_2023} formulated their solution as an integer linear programming (ILP) problem, where they associate RWID with tracklets under the assumption that no identity switch occurs within a tracklet. Their approach accounts for RWID uncertainty by assigning different weights to each RWID-tracklet pair. Since ILP is NP-hard \citep{aloul2002generic}, \citep{Camilleri_2023} opted to group frames sharing the same subset of active tracklets, transforming the tracking problem into a set-covering problem. While this problem remains NP-hard, frame grouping improves the method’s time efficiency. However, if only a small number of frames can be grouped, the algorithm becomes slower. Consequently, in certain tracking scenarios, the method may experience significant delays, especially when large frame groups cannot be formed. Moreover, it assumes that individual tracklets are perfectly reliable, which is not always the case especially for large groups of frames. Lastly, RWID identifications must be frequent and in some contexts, such as markerless livestock tracking, identifications are sparser, making the approach less applicable.

 \textbf{Contribution}: In this paper, we propose a long-term identity-aware MOT framework that is able to use sparse and uncertain RWID information from a feeder to improve tracking of a base tracker. The framework is based on the formulation of the tracking problem as a Hidden Markov Model (HMM) problem \citep{Degirmenci2014,SIN199531}. Formulating the problem like this provides a new identity-aware framework with a solution in $O(TN^3)$ (where $T$ is the number of frames and $N$ the number of RWIDs). The problem formulation provides the ability to combine classic MOT approaches and uncertain sparse identifications. Thus, by using our new HMM framework, approaches like \citep{Yu2016, 7298718} could be used as RWID identity sources and combined with other MOT approaches like \citep{Luo2021, Zhang2021, bewley2016simple, Wojke2018, 10030267, Bastani2021, Vondrick2018}.

In addition to the HMM framework, we also provide a new long-term 10-minute tracking dataset of pigs with 21 uncertain RWID identifications obtained from a feeder station.

 This approach is the first of its kind in the livestock industry. We tested the framework on the pig dataset with ByteTrack \citep{Zhang2021} as the base tracker and compared it to ByteTrack and ByteTrack with re-identification. The use of our HMM framework improved the identification F1 score with a minimal number of identifications. Our framework also exhibits stable performance over time unlike the other approaches whose performance deteriorate over time. It also shows a faster convergence in terms of the number of uncertain identifications compared to ByteTrack with re-identification, which was hindered by incorrect identifications. Those findings were validated on six videos from the MOT17  \citep{dendorfer2021motchallenge} dataset, on which the approach was tested due to the lack of open tracking datasets similar to the one we provide. On MOT17, both ByteTrack and FairMOT \citep{zhang2021fairmot} were tested in combination with our HMM framework and we found similar results. Overall, our new HMM framework could provide RWID to objects in a video and improve MOT when sparse and uncertain RWID are provided by an identification source in a scene.
 
\section{Method}

\subsection{HMM Framework}
\subsubsection{Overview}

\begin{figure*}[ht!]
\begin{center}
\includegraphics[width=17cm]{ 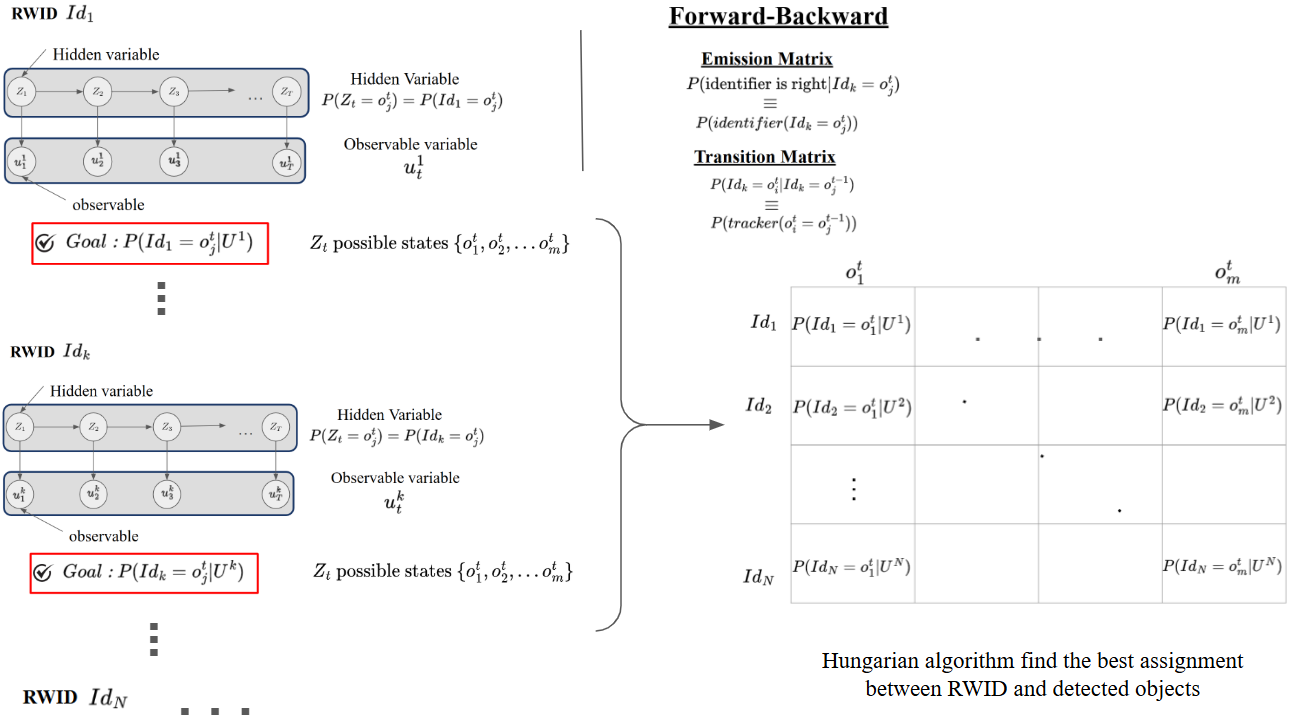}
\caption{
Our framework defines one HMM for each real-world identity (RWID) of interest (left part of the figure). For each HMM, the hidden state represents the probability of matching the RWID with a detected object (eg. a pig), while the observable state corresponds to the correctness of the $identifier$'s decisions. A forward-backward algorithm is applied to estimate the probability of matching an RWID to detected objects given all identifications made by the $identifier$ of that RWID. Finally, the Hungarian algorithm is used to find the optimal assignment between all RWIDs and detected objects at each frame.}

\label{fig:Framework_explained}
\end{center}
\end{figure*}

Let us consider the assignment of RWIDs to detected objects using a combination of multiple HMMs (i.e. one HMM for each RWID) \citep{Degirmenci2014,SIN199531}. Each HMM is responsible for assigning one given RWID to detected objects (eg. pigs), by taking into account all identifications of that RWID by the $identifier$ as presented in figure \ref{fig:Framework_explained}

The goal is to determine the optimal assignment of RWIDs to detected objects while considering all identifications made by the $identifier$. In our HMM formulation, the RWIDs obtained from the $identifier$ serve as observations. Given this formulation, the forward-backward algorithm can efficiently estimate the probability of assigning each RWID to an object at every frame in polynomial time \citep{Degirmenci2014}. 
We chose not to use Viterbi because it enforces a globally optimal RWID assignment over the entire video without modeling interactions between RWIDs, which means per-frame adjustments cannot be made. Instead, we combine the forward-backward algorithm with the Hungarian algorithm to merge assignments from individual HMMs, ensuring a more coherent and comprehensive tracking solution in polynomial time. The complexity of the forward-backward algorithm is in $O(Tm^2)$ for the individual HMM and in $O(NTm^2)$ for all RWIDs. Since the information from the individual HMM is integrated using an Hungarian algorithm at every frame the complexity of this step is in  $O(TN^3)$. Since $m \leq N$, we can assume the overall complexity of the approach to be in $O(TN^3)$.

Applying the forward-backward algorithm requires transition and emission matrices, which can be obtained from the $tracker$ (e.g., ByteTrack or FairMOT) and the $identifier$ (e.g., a feeding station that probabilistically assigns RWIDs to nearby pigs), respectively. In Section \ref{sec:problem_formulation}, we will discuss in detail how the $tracker$ and $identifier$ generate these matrices.

\subsubsection{Problem formulation}

\label{sec:problem_formulation}
\textbf{Notations.}

Let us assume:
\begin{itemize}
    \item  $T$ the total number of frames in a video. 
\item $O_t=\{o^t_1,...,o^t_{m}\}$ the set of $m$ objects $o$ detected at frame $t$. 
\item $ID =\{Id_1,...,Id_k,..,Id_N\}$  the set of all $N$ RWID. 
\item $P(Id_k = o^t_j)$  the probability that RWID $Id_k$ is the real identity of the detected object $o^t_j$ at frame $t$.
\item $P(Identifier (Id_k = o^t_j))$  the probability that $o^t_j$ has the identity $Id_k$ at time $t$ according to the $identifier$.
\item $U^k=\{u^{k}_{1}, ..., u^{k}_T\}$ is the set of observations on identity $Id_k$ (by the $identifier$). Here $u^k_t$ is the event that the $identifier$ choice is right (the $identifier$ identified well the object with identity $Id_k$). So we have $P(u^k_t | Id_k =o^t_j) = P(Identifier (Id_k =o^t_j))$.

Unfortunately, in some cases, identifications are highly sparse and not available at every time step $t$. To address this sparsity, we assume that when no identification is available, the $identifier$ assigns an equal probability to all objects using $P(Identifier (Id_k = o^t_j)) = \frac{1}{m} \forall o^t_j \in O_t$.

\item $P( tracker(o^{t-1}_i= o^{t}_j))$ is the probability that $o^{t-1}_i$ matches $o^{t}_j$ in two consecutive frames (i.e. they are the same object). This information is provided by the $tracker$.
\end{itemize}

The goal of our HMM formulation is to obtain $P(Id_k = o^t_j | U^k)$ (i.e. the probability that the RWID $Id_k$ is really associated to $o^t_j$ knowing the $identifier$ observations) for each $t,j,k$. By considering a HMM for each $k$ having the $Id_k$ state, and each $o^t_j \in O_t$ as possible values at time $t$, it is possible to compute $P(Id_k = o^t_j | U^k)$ using the forward-backward algorithm. However, the use of the forward-backward algorithm requires the transition and emission matrix at each $t$ providing respectively equation \ref{eq:transition_matrix} and \ref{eq:emission_matrix}: 

\begin{equation}
{\scriptstyle
    P(Id_k = o^t_j| Id_k = o^{t-1}_l) = P(tracker(o^{t-1}_l= o^{t}_j)) 
}
    \label{eq:transition_matrix}
\end{equation}

$ \forall o^{t-1}_l \in O_{t-1}$  and $ o^t_j \in O_t$ which is given by the \textit{tracker}.
\begin{equation}
    P(u^k_t | Id_k=o^t_j) = P(Identifier (Id_k =o^t_j))
    \label{eq:emission_matrix}
\end{equation}
$ \forall o^t_j \in O_t.$ which is given by the \textit{identifier}.

\begin{figure*}[ht!]
\begin{center}
\includegraphics[width=14cm]{ 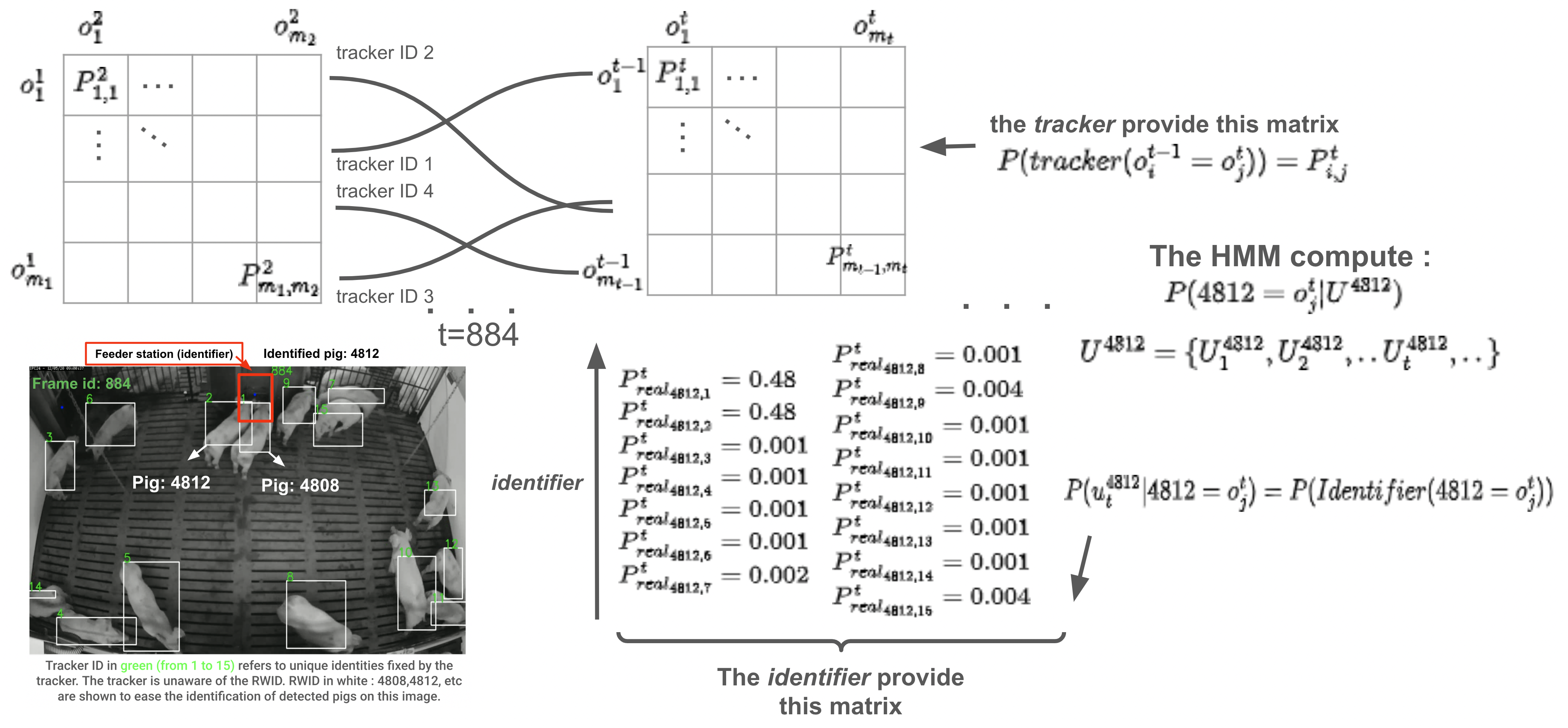}
\caption{HMM tracking reformulation explained. (Bottom left) Here, at frame number t=884, the pig 4812 is identified at the feeder, but there are two pigs near the feeder. In this situation, pigs with tracker IDs 1 and 2 will have the highest probabilities of being assigned to RWID 4812 since they are close to the feeder. We retrieve this set of probabilities as a line of the emission matrix at frame number 884. The process is similar for all frames. At the top we could see the transition matrix provided by the tracker at each frame number. These sets of emissions and transition matrices will be used together in the HMM framework to propose probabilities of assigning a RWID to a detected object given all identifications at the feeder. In this figure, $p^t_{real_{k,j}}$ = $p(Identifier (Id_k =o^t_j))$.}
\label{fig:hmmexplained}
\end{center}
\end{figure*}

\subsubsection{Problem-solving: the forward-backward algorithm for our HMM}

With the forward-backward algorithm we can process $\alpha_j(t)=P(u^k_1, ..., u^k_t, Id_k = o^t_j )$, and 
 $\beta_j(t)$= $P(u^k_{t+1},..., u^k_T | Id_k = o^t_j )$ as in equation \ref{eq:alpha} and \ref{eq:beta} respectively and obtain the marginal probability of a given state \citep{18626}.

\begin{multline}
   \alpha_j(t)=\sum^{|O_ {t-1}|}_{l=1}\alpha_l(t-1) \times P(Identifier (Id_k =o^t_j)) \\  \times P( tracker(o^{t-1}_l= o^{t}_j))
    \label{eq:alpha} 
\end{multline}
with $\alpha_j(0)=1 \quad \forall j$.
\begin{multline}
    \beta_j(t)=\sum^{|O_ {t+1}|}_{l=1} \beta_l(t+1) \times P(Identifier(Id_k = o^{t+1}_l))) \\ \times P(tracker( o^{t+1}_l = o^{t}_j))
    \label{eq:beta} 
\end{multline}
with $\beta_j(T)=1 \quad  \forall j$.

We are interested in finding $P(Id_k = o^t_j | U^k)$. With the forward-backward algorithm, it is possible to process 
$P(Id_k = o^t_j | U^k)$ as $\frac{P(U^k, Id_k = o^t_j)}{P(U^k)} = \frac{\alpha_j(t)\beta_j(t)}{P(U^k)} $ \citep{18626}. However $P(U^k)$ is a constant value over time accross detections. Knowing that we are interested in finding $o^t_j$ with the highest probability of having identity $Id_k$, performing the assignment by considering $\alpha_j(t)\beta_j(t)$ as matching value between $o^t_j$ and $Id_k$ will provide the same result as considering $P(Id_k = o^t_j | U^k)$.

\subsubsection{Numerical stability}

In the context of long-term tracking, the HMM is applied on a large numbers of frames. Since $\alpha$ and $\beta$ are estimated from successive multiplication of small probabilities $\leq 1$, this could rapidly lead to numerical instability \citep{bishop2006pattern}. To solve this issue, for each $t$, we normalize $\alpha(t)$ and $\beta(t)$ by $\sum_j \alpha_j(t)$ and $\sum_j \beta_j(t)$. Instead of using $\alpha(t)$ and $\beta(t)$ on the forward-backward algorithm, we use $\hat{\alpha}_j(t) = \frac{\alpha_j(t)}{\sum_j \alpha_j(t)}$ and $\hat{\beta}_j(t) = \frac{\beta_j(t)}{\sum_j \beta_j(t)}$. By this process, the probability propagation is maintained and we avoid going under machine epsilon. However, by doing so, the exact probability $P(Id_k = o^t_j | U^k$) is lost but for each $t$ and $j$, according to \citep{bishop2006pattern} there exists $c^t$ such that $\hat{P}(Id_k = o^t_j | U^k)= P(Id_k = o^t_j | U^k) \times c^t$. Therefore they can be used equivalently at each $t$.

\subsection{Dataset}
\subsubsection{Experimental setup}

The dataset was collected from a pen of growing pigs of 12.7 $m^2$ $(4.87m \times 2.6m)$ shortly after they were transferred from the nursery. A camera (4MP resolution with a $2.7 \sim 13.5 \, \text{mm varifocal lens}$) recording at 25 fps was mounted on the ceiling to capture the entire area. A feeding station (i.e. the $identifier$) is positioned in the top part of the pen (Figure \ref{fig:station})

\subsubsection{Animal subjects}

\begin{figure*}[ht!]
\begin{center}
\includegraphics[width=14cm]{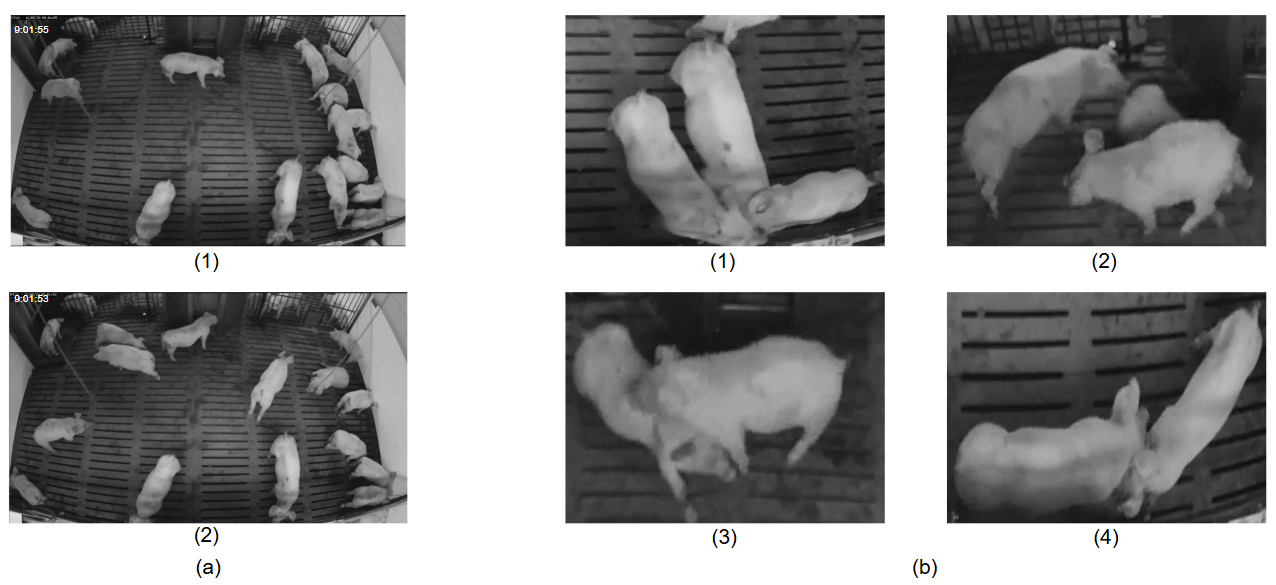 }
\caption{Snapshots of pig activities in the park, extracted from our 10-minute video, are presented. (a) Two distinct frames captured 2 seconds apart ((a-1) at 09:01:53 and (a-2) at 09:01:55) highlight the changes in the pigs' positions over this short interval, illustrating their high levels of activity. (b) Various behaviors are depicted, including pigs moving and interacting with each other. These activities often result in occlusions and erratic movements, compounded by the fact that all pigs share similar appearances. The combination of high activity levels and visual similarity among the pigs significantly increases the complexity of tracking in this scenario.}
\label{fig:activities}
\end{center}
\end{figure*}

In the pen, there was a fixed number of 15 pigs of similar age, weight, and appearance, as is common in the livestock sector. The behavior of the 15 pigs was recorded during 10 minutes from 9:00 to 9:10 am (Eastern Time). During this 10-minute video, pigs were energetic and displayed various behaviors, including feeding (drinking and eating), playing, running across the pen, and engaging in interactions with other pigs including minor aggressions and play (Figure \ref{fig:activities}).

\subsubsection{Data annotation}

The goal of the dataset was to provide a comprehensive tracking resource suitable for various types of trackers, including those that perform tracking through pose estimation. For each pig, key points (the two ears, tail, neck, and nose) were manually annotated along with their identities. Based on these key points, a bounding box was created as the smallest enclosing rectangle around them. In certain situations, when the pig was bent with its back facing the camera, only the neck and tail were visible. In these cases, a rectangle was constructed, centered on the line formed by the two points with a fixed width corresponding to the average width of all pig bounding boxes.

The 10-minute video was manually annotated with the key points and identities of the 15 animals, approximately every second, resulting in 780 annotated frames. In addition to these tracking annotations, 21 sporadic identifications from the feeder are provided, distributed throughout the video as shown in Figure \ref{fig:F1_comp}. The 21 identifications were recorded from 11 pigs that passed by the feeder. The full dataset and annotations are available at :  \href{https://github.com/ngobibibnbe/uncertain-identity-aware-tracking/tree/main/dataset}{https://github.com/ngobibibnbe/uncertain-identity-aware-tracking/tree/main/dataset}.

\section{Experiments}
\subsection{Testing the HMM framework on a new 10-minute video}

To test our framework, we used  ByteTrack \citep{Zhang2021} as the base tracker. We evaluated three different tracking approaches in our experiments on the 10-minute video: ByteTrack alone (ByteTrack), ByteTrack with re-identification (ByteTrack+Re-ID), and ByteTrack in combination with our HMM framework (ByteTrack+HMM).

ByteTrack is a tracking-by-detection model and it requires a detection model. We trained a YOLOX \citep{Ge2021} model from 11700 annotations of pigs (80\% training, 20\% validation) using 500 epochs and got an mAP of 96.5\% for the training set and 95\% for the validation set.

For the ByteTrack+HMM approach, ByteTrack provides the transition matrix. As for the emission matrix, when an RWID $Id_k$ is detected at a given frame $t$, the probability, according to the $identifier$, that $Id_k$ corresponds to a detected animal $o_j$ at time $t$ is computed as a function of the pig's distance to the feeding station. The closer the pig is to the station, the higher the probability : \[ P(\text{Identifier}(Id_k=o^t_j))= \frac{1/\text{distance}(o^t_j, \text{station}) }{\sum_l 1/\text{distance}(o^t_l, \text{station}) }. \]
 When no identification is available at a given time $t$, the identification probability is set to :
 \[
P(\text{Identifier}(Id_k=o^t_j))=\frac{1}{N}=\frac{1}{15},
\]
since there is no indication of which detected animal could be associated with a given RWID. Here, 15 corresponds to the total number of pigs in the video. From this, individual HMMs are constructed for the RWIDs identified at least once at the feeder, and matching probabilities between detected animals and RWIDs are computed using the forward-backward algorithm. The Hungarian algorithm is then used to determine the optimal assignment between RWIDs and detected animals at each frame $t$. In each assignment, if the matching probability between an RWID  $Id$ and a detected animal $o$ is lower than $1/N$, the match is disregarded due to high uncertainty. As a result, the animal is considered undetected rather than misidentified.

For the ByteTrack and ByteTrack+Re-ID approaches, ByteTrack’s internal object IDs are matched with the closest RWIDs at the very first frame. For the ByteTrack approach the RWIDs are associated to the same internal object throughout the entire video. For the ByteTrack+Re-ID approach, if a pig is detected at the feeder station but does not have the expected RWID, an identity switch is assumed and the identities of the detected object at the feeder and the object that previously held that RWID are swapped. This correction is then applied to future frames.

Since we have access to real RWIDs and their corresponding pig bounding boxes as ground truth approximately every second throughout the 10-minute video, we can effectively evaluate the performance of different approaches by comparing the RWIDs assigned by each approach to those in our dataset.

As noted by \cite{7780789, Camilleri_2023}, traditional MOT metrics do not fully address the requirements of identity-aware tracking, as they do not account for the correct assignment of RWIDs. Therefore, new metrics are needed that consider both the detection of ground truth objects and the accurate assignment of RWIDs.

To evaluate the performance of different approaches, we used I-TP (Identity-aware True Positive: ground truth object detected with the correct RWID), I-FP (Identity-aware False Positive: either an unmatched detected object OR a ground truth object detected with the wrong RWID), and I-FN (Identity-aware False Negative: either a ground truth object not detected OR a ground truth object detected without an assigned RWID). Additionally, we computed Micro-Precision, Micro-Recall, and Micro-F1 Score, following the methodology in \cite{7780789}. These metrics were chosen because they provide a comprehensive measure of how well each approach assigns the correct RWID to detected individuals throughout the video.

Unlike \cite{7780789}, which applies a fixed 1-meter cutoff, we consider a detected object matched to a ground truth object if $\text{IOU} > 0$. Based on this criterion, Micro-Precision, Micro-Recall, and Micro-F1 Score are computed as follows:
\begin{align*}
\text{Micro-Precision} &= \frac{\text{I-TP}}{\text{I-TP} + \text{I-FP}} \\
\text{Micro-Recall} &= \frac{\text{I-TP}}{\text{I-TP} + \text{I-FN}} \\
\text{Micro-F1 score} &= \frac{2 \cdot \text{Micro-Precision} \cdot \text{Micro-Recall}}{\text{Micro-Precision} + \text{Micro-Recall}}
\end{align*}

\begin{table*}[ht!]
\centering\begin{tabular}{|c|c|c|c|c|c|c|c|c|c|c|} 
 \hline
Dataset & I-TP & I-FP & I-FN   & Micro-Precision & Micro-Recall & Micro-F1 score  \\ 
 \hline
{ByteTrack}  & 4125 & 4252 & \textbf{291} & 0.49 & 0.48 & 0.48   \\ 
 \hline
{ByteTrack+Re-ID} & 4222 & 4093  & \textbf{291} & 0.51 & 0.49 & 0.50  \\ 
 \hline
{ByteTrack+HMM }  & \textbf{5264 } & \textbf{1751 } & 1642  & \textbf{0.75}  & \textbf{0.61}  & \textbf{0.67}  \\ 
 \hline 
\end{tabular}
\caption{Tracking performance of the different approaches using Identity-aware True Positive (I-TP), Identity-aware False Positive (I-FP), Identity-aware False Negative (I-FN), Micro-Precision, Micro-Recall, and Micro-F1 score (see the text for their definitions).}

\label{table_compar}
\end{table*}

\begin{figure}[h!]
\centering
\includegraphics[width=8.5cm]{ 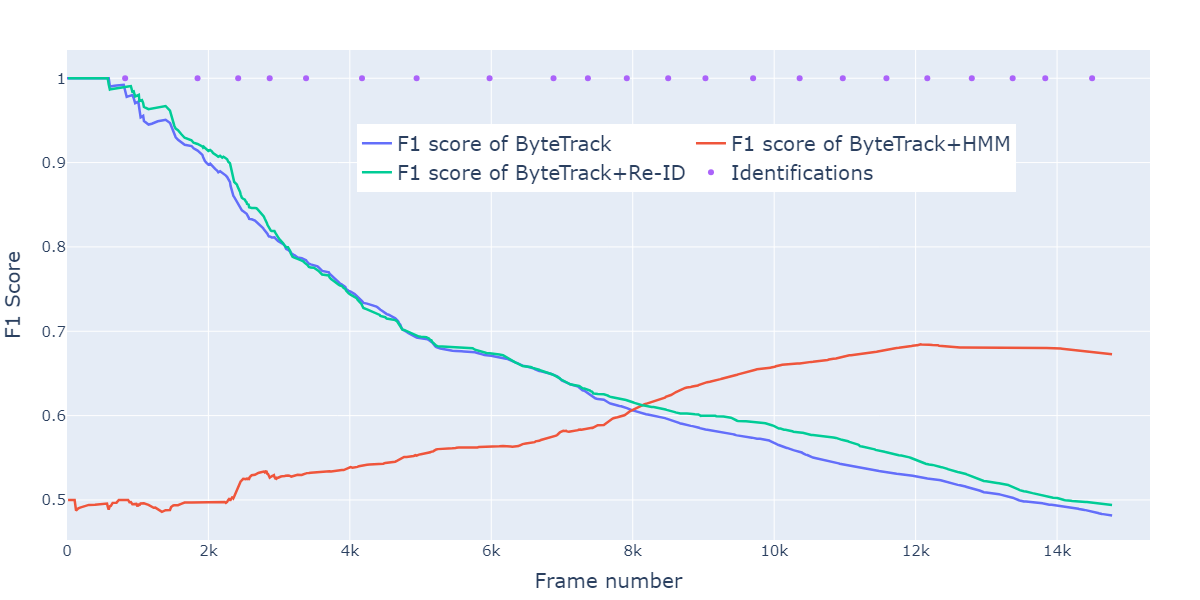}
   \caption{F1 score over time.   }
\label{fig:F1_comp}
\end{figure}

As shown in Table \ref{table_compar}, our new HMM framework improves identity-aware tracking in comparison to ByteTrack alone and ByteTrack+Re-ID for most metrics. We notice higher I-TP, lower I-FN, higher Micro- Precision, Recall, and F1 score. This improvement can be attributed to the HMM framework's ability to propagate identifications effectively, as well as its probabilistic nature, which facilitates the recognition of low-probability assignments. However, this probabilistic approach also accounts for the observed increase in I-FN, as it inherently considers a broader range of potential identifications, including those with lower confidence.

When analyzing the evolution of the performance of the three approaches over time as a function of an increasing number of identifications (Figure \ref{fig:F1_comp}), we can observe that the performance of ByteTrack+HMM exhibits greater stability compared to ByteTrack alone or ByteTrack+Re-ID, both of which show a decline in performance as time progresses. Additionally, we can see that as the number of identifications increases (after 6K frames), the HMM approach becomes more precise.

\subsection{Testing the impact of uncertainty and the number of identifications on the HMM framework}
\label{testing_over_identifications}
To assess the performance of ByteTrack+Re-ID and ByteTrack+HMM as a function of the level of uncertain identification information, and their respective robustness to identification errors, we generated artificial identifications over the annotated 10-minute video. Here, 25\% of the artificial identification information was randomly assigned to all animals, while 75\% was assigned with higher probability to the animal with the real identity based on the distance to the ground truth from which we extract the identity. To ensure that our random simulations are representative this procedure was repeated 20 times.

\begin{figure}[h!]
\centering
\includegraphics[width=8cm]
{ 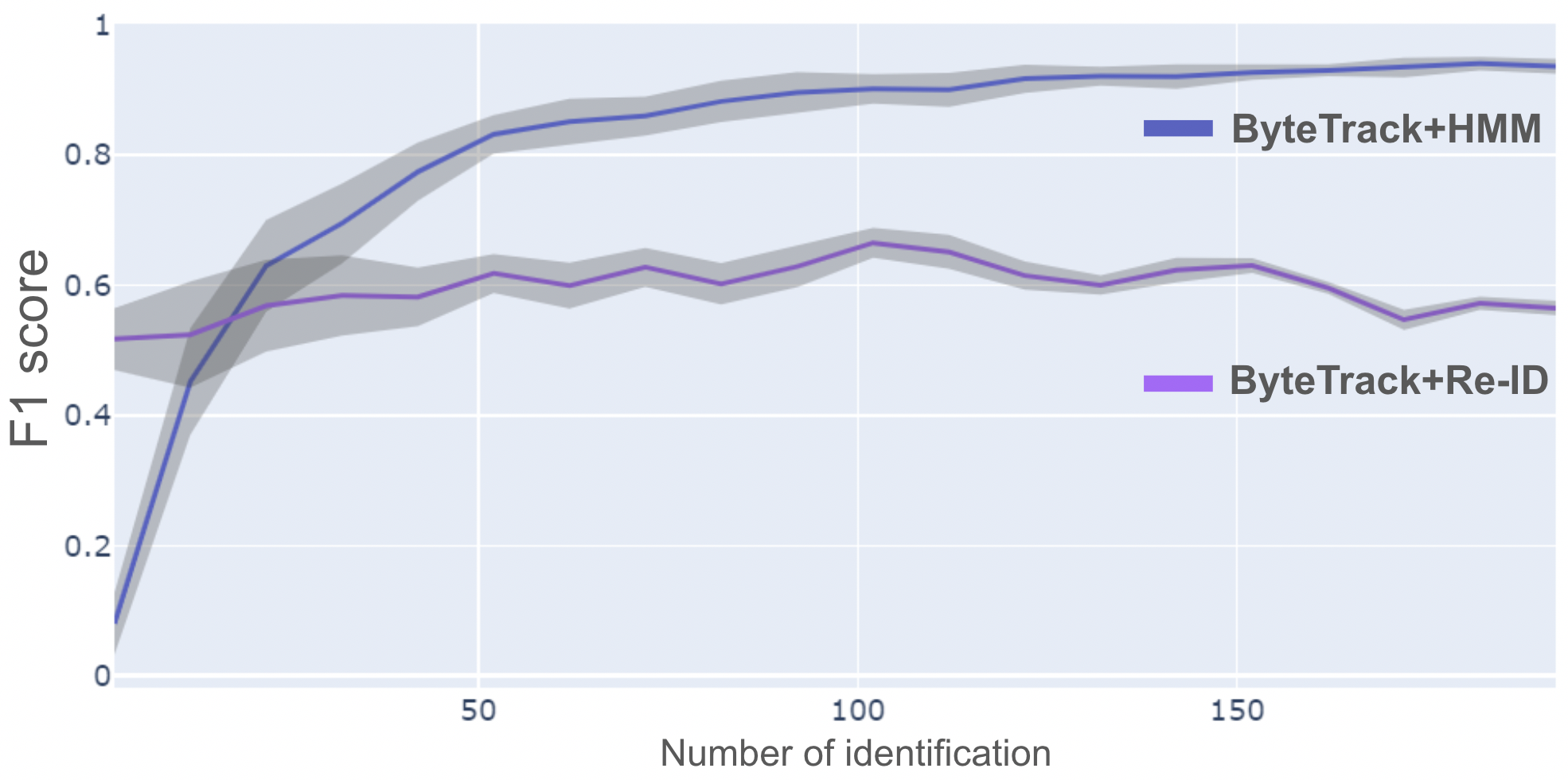}
   \caption{F1 score as a function of the number of identifications from the $identifier$. Error bars represent standard deviation from 20 random simulations where 25\% of observations have been shuffled.}
\label{fig:evol}
\end{figure}

On Figure \ref{fig:evol} we see that ByteTrack+HMM tends to have higher F1 score compared to ByteTrack+Re-ID when the number of identifications is over 8. This is due to the fact that the ByteTrack+Re-ID model obtain all the RWID information at the beginning of the video while ByteTrack+HMM needs to obtain them from the identifications. For RWID that are never identified by the $identifier$ (e.g. pigs not visiting the feeder in a given video), the matching probabilities between all detected objects and those RWID will be the same. For those animals, this will lead to lower performance for ByteTrack+HMM.

\subsection{Testing the HMM framework with another MOT approach (FairMOT) on MOT17 and MOT20 }

To further test our HMM framework in videos outside the livestock sector, we used videos of  MOT17 and MOT20  benchmarks \citep{dendorfer2021motchallenge}. These datasets feature videos of pedestrians filmed from different points of view (e.g. moving camera, elevated viewpoint, nighttime, daytime, etc.). The videos are relatively short, ranging from 18 seconds to 1 minute, with frame rates varying from 14 to 30 fps for MOT17. For MOT20 videos are still short ranging from 17 seconds to 2 minutes and 13 seconds, with frame rates of 25 fps. The number of people also varies between videos, from 26 to 133 in total for MO17, and from 90 to 1211 for MOT20. In all those videos, there are many individuals entering and exiting the scene throughout the video. The MOT17 and MOT20 dataset does not provide RWID identifications so we decided to simulate artificial identifications based on the ground truth provided in the MOT17 dataset. This is similar to what was presented in the previous section, but for the MOT17 and MOT20 videos.

We were also interested to test if the performance of our new HMM framework could also be extended to another MOT approach. For this experiment we tested both ByteTrack and FairMOT. We evaluated the performance of ByteTrack and FairMOT combined with Re-ID or our HMM framework as a function of the number of artificial random identifications evenly distributed throughout the videos. For our MOT17 and MOT20 experiment, we used the object detection provided by the model provided on the github of the authors \citep{Zhang2021,zhang2021fairmot} for both ByteTrack and FairMOT respectively. We conducted the tests on the validation sets of the benchmarks, which represent the second half of each video.

\begin{figure*}[htbp!]
    \centering
\begin{subfigure}[b]{0.48\textwidth}
        \centering
        \includegraphics[width=7.5cm ,  height=4.5cm]{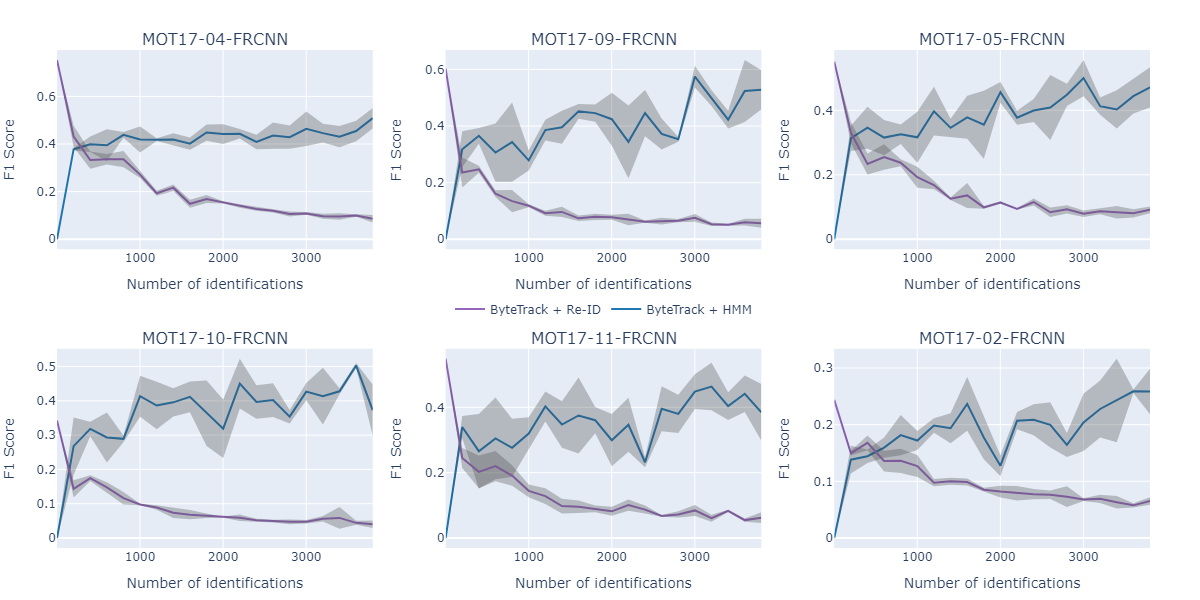}
        \caption{ByteTrack as base tracker with 25\% random observations}
        \label{fig:bytetrack_random}
    \end{subfigure}
    \hfill
        \begin{subfigure}[b]{0.48\textwidth}
        \centering
        \includegraphics[width=7.5cm ,  height=4.5cm]{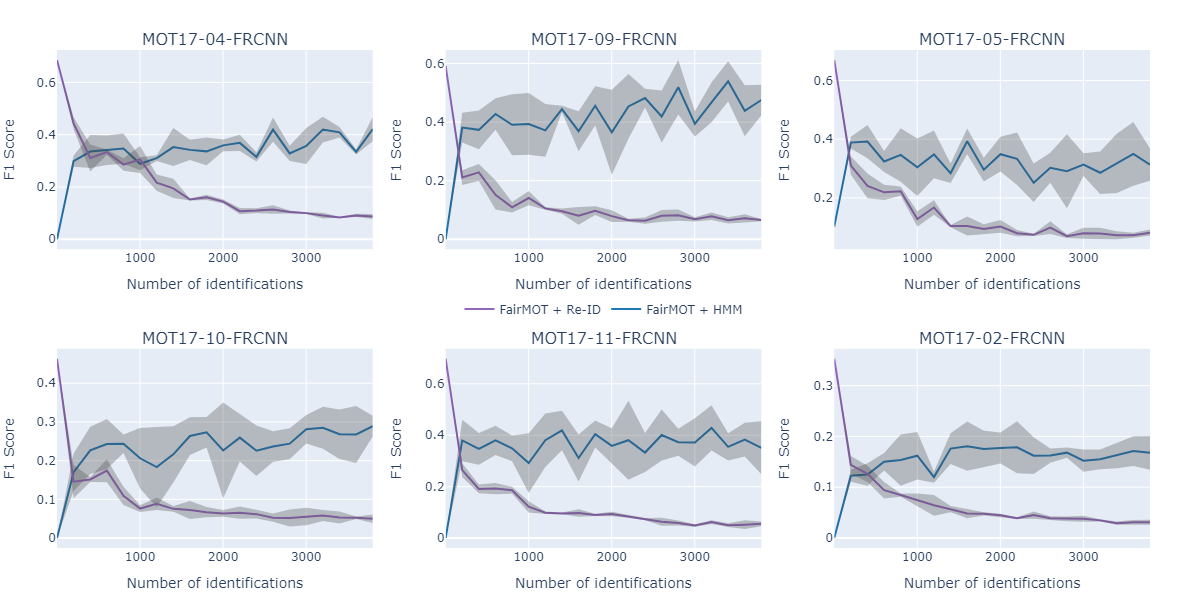}
        \caption{FairMOT as base tracker with 25\% random observations}
        \label{fig:fairmot_random}
    \end{subfigure}

 \vskip\baselineskip
    \begin{subfigure}[b]{0.48\textwidth}
        \centering
        \includegraphics[width=\textwidth]{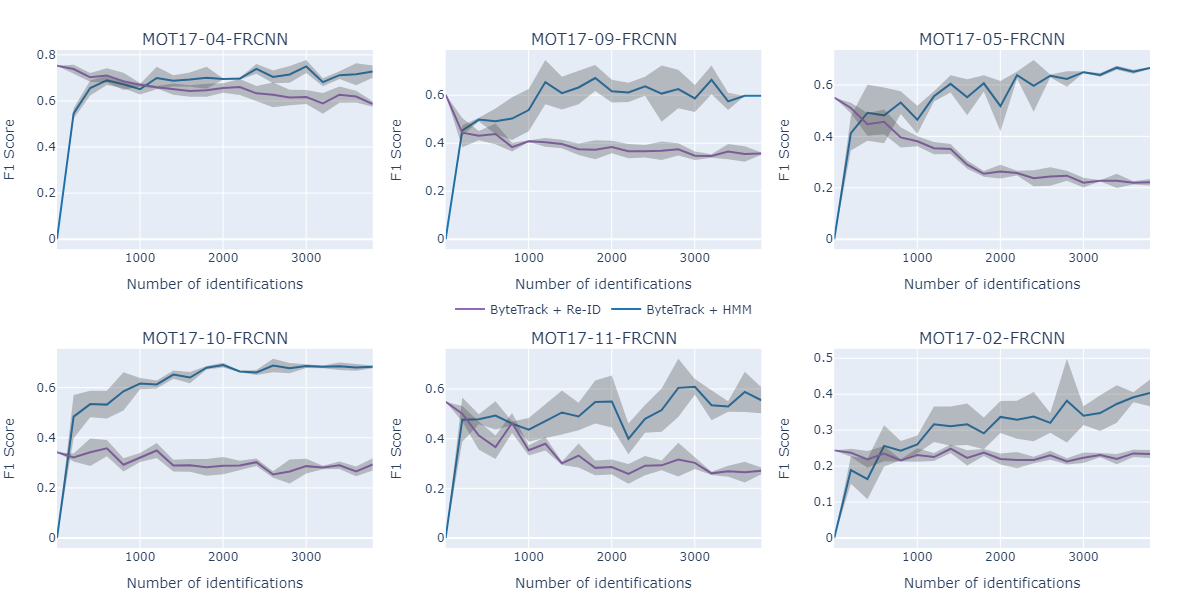}
        \caption{ByteTrack as base tracker with no random observations
        }
        \label{fig:bytetrack_no_random}
    \end{subfigure}
    \hfill
    \begin{subfigure}[b]{0.48\textwidth}
        \centering
        \includegraphics[width=\textwidth]{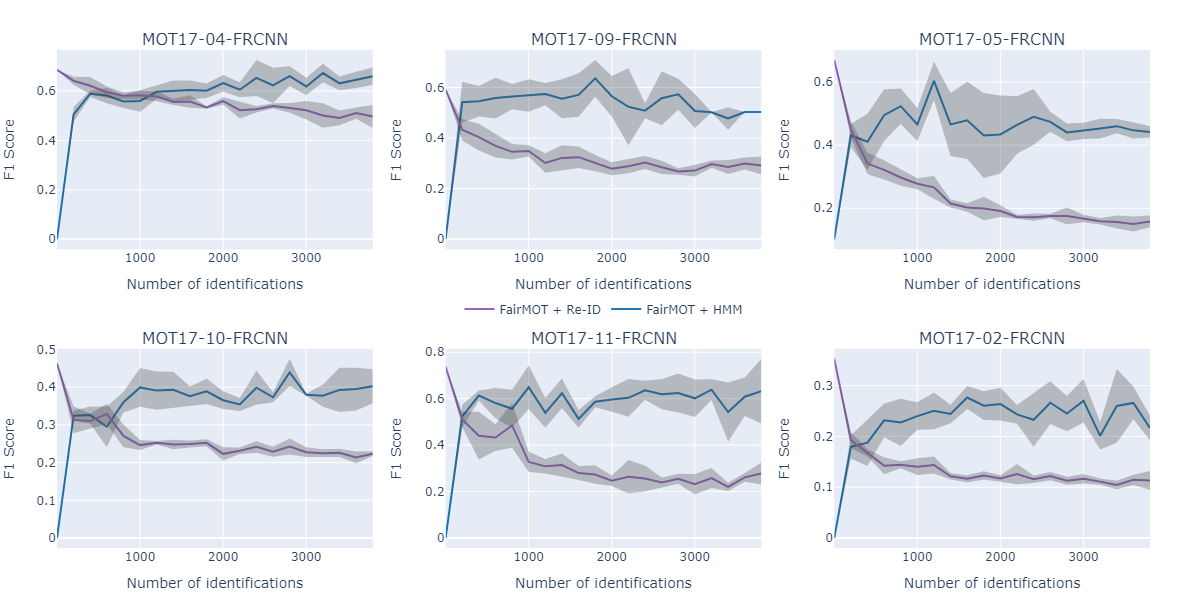}
        \caption{FairMOT as base tracker with no random observations }
        \label{fig:fairmot_no_random}
    \end{subfigure}

    \vskip\baselineskip
    \begin{subfigure}[b]{0.48\textwidth}
        \centering
            \includegraphics[width=\textwidth]{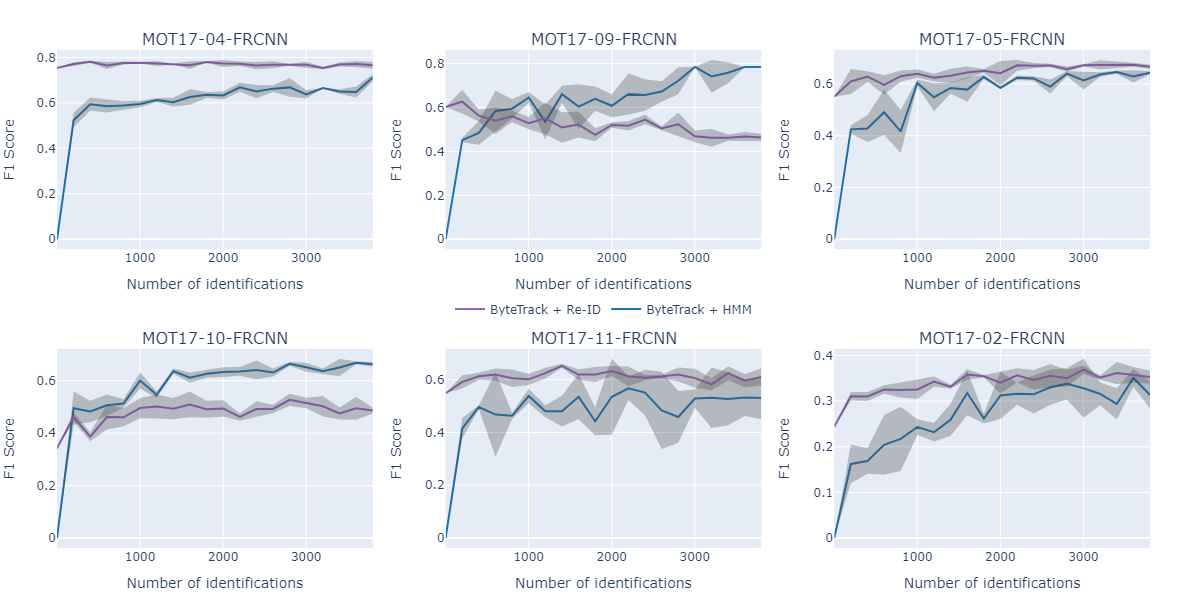}
        \caption{ByteTrack as base tracker with filtered identifications only
        }
        \label{fig:bytetrack_no_random_filtered}
    \end{subfigure}
    \hfill
    \begin{subfigure}[b]{0.48\textwidth}
        \centering
        \includegraphics[width=\textwidth]{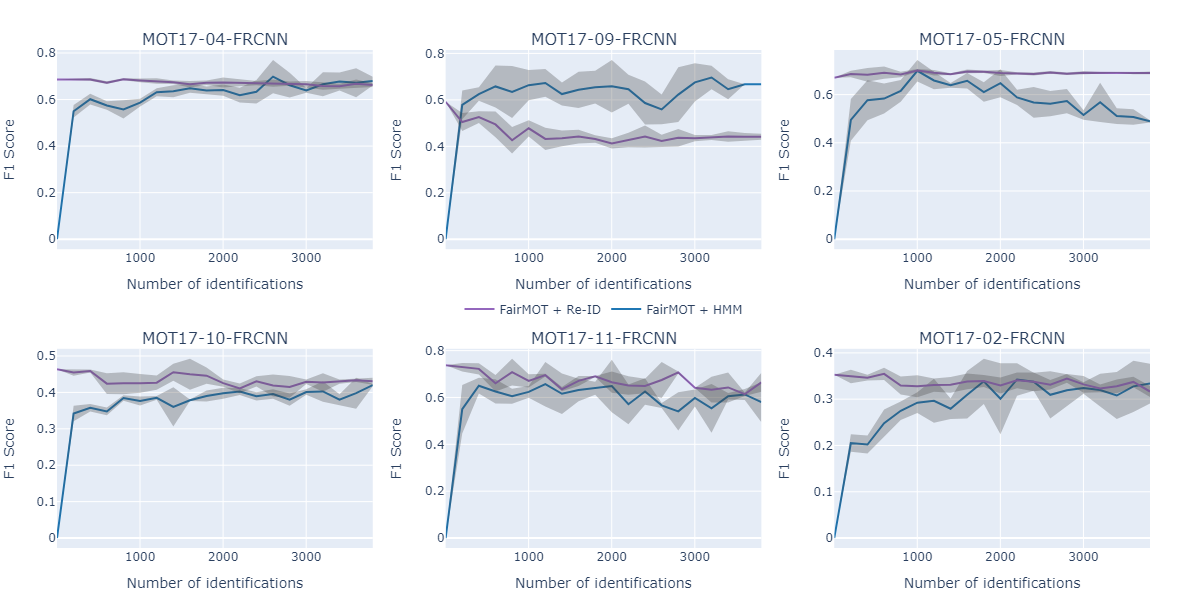}
        \caption{FairMOT as base tracker with filtered identifications only }
        \label{fig:fairmot_no_random_filtered}
    \end{subfigure}

\caption{F1 score as a function of the number of identifications from the artificial identifier in the MOT17 dataset. Error bars present standard deviation from 20 random simulations.  In \ref{fig:bytetrack_random} and \ref{fig:fairmot_random}, observations are generated such that 25\% of them are randomly distributed. These do not include identification errors caused by crowded scenes during the artificial observation generation process, so there are more than 25\% of misleading observations generated. In \ref{fig:bytetrack_no_random}  and   \ref{fig:fairmot_no_random} none of the observations are randomly distributed, but rather are all intended to be correct. In \ref{fig:bytetrack_no_random_filtered} and  \ref{fig:fairmot_no_random_filtered}, observations are filtered to remove misleading identification candidates, and no random identifications are added to limit the number of misleading observations.}

\label{fig:bytetrackmot17}
\end{figure*}

\begin{figure*}[htbp!]
    \centering
\begin{subfigure}[b]{0.48\textwidth}
        \centering
        \includegraphics[width=7.5cm ,  height=4.5cm]{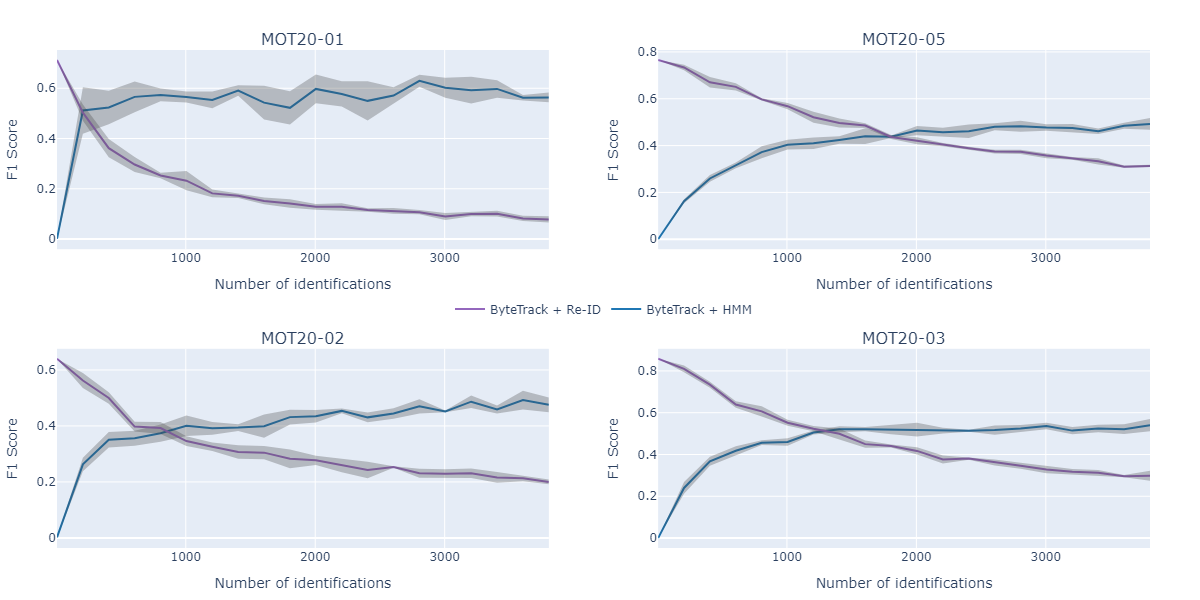}
        \caption{ByteTrack as base tracker with 25\% random observations}
        \label{fig:bytetrack_random_mot20}
    \end{subfigure}
    \hfill
        \begin{subfigure}[b]{0.48\textwidth}
        \centering
        \includegraphics[width=7.5cm ,  height=4.5cm]{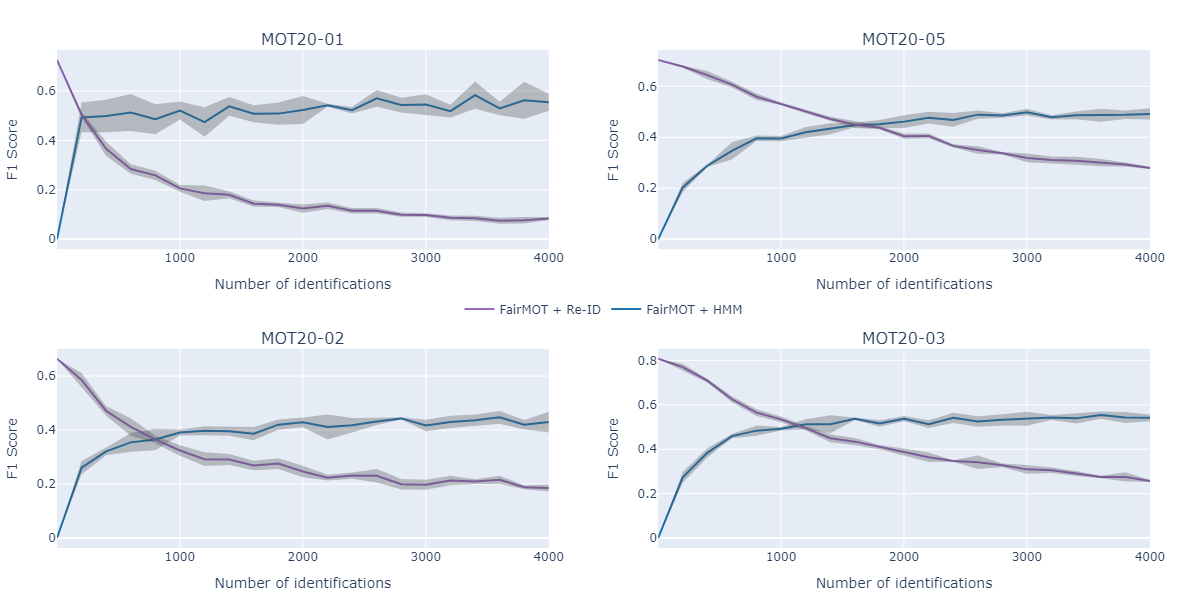}
        \caption{FairMOT as base tracker with 25\% random observations}
        \label{fig:fairmot_random_mot20}
    \end{subfigure}

 \vskip\baselineskip
    \begin{subfigure}[b]{0.48\textwidth}
        \centering
        \includegraphics[width=\textwidth]{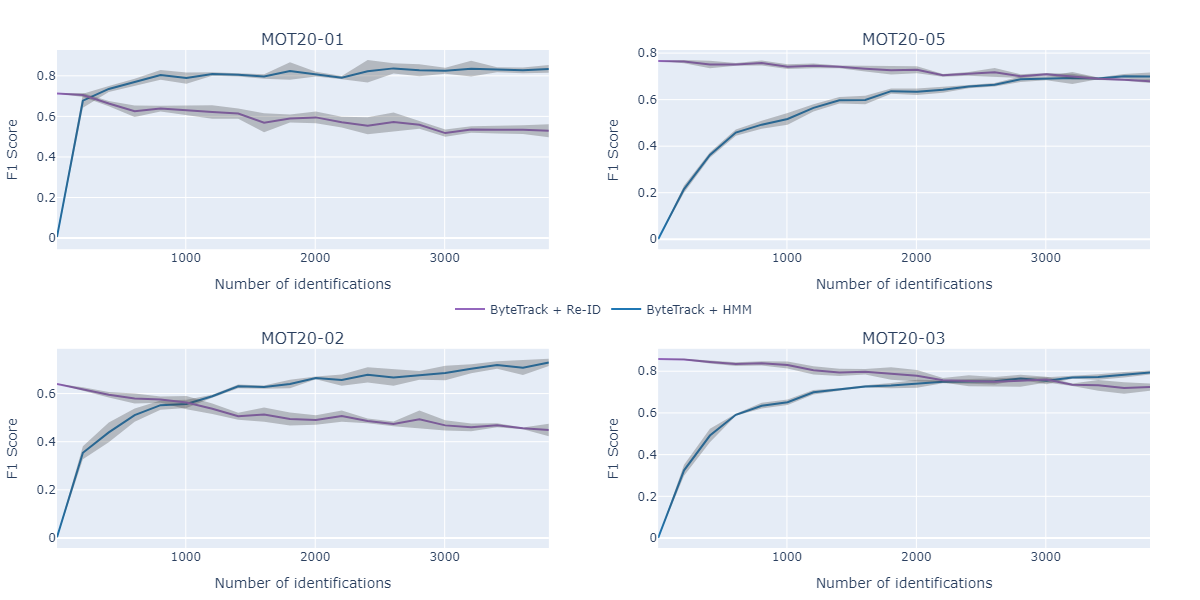}
        \caption{ByteTrack as base tracker with no random observations
        }
        \label{fig:bytetrack_no_random_mot20}
    \end{subfigure}
    \hfill
    \begin{subfigure}[b]{0.48\textwidth}
        \centering
        \includegraphics[width=\textwidth]{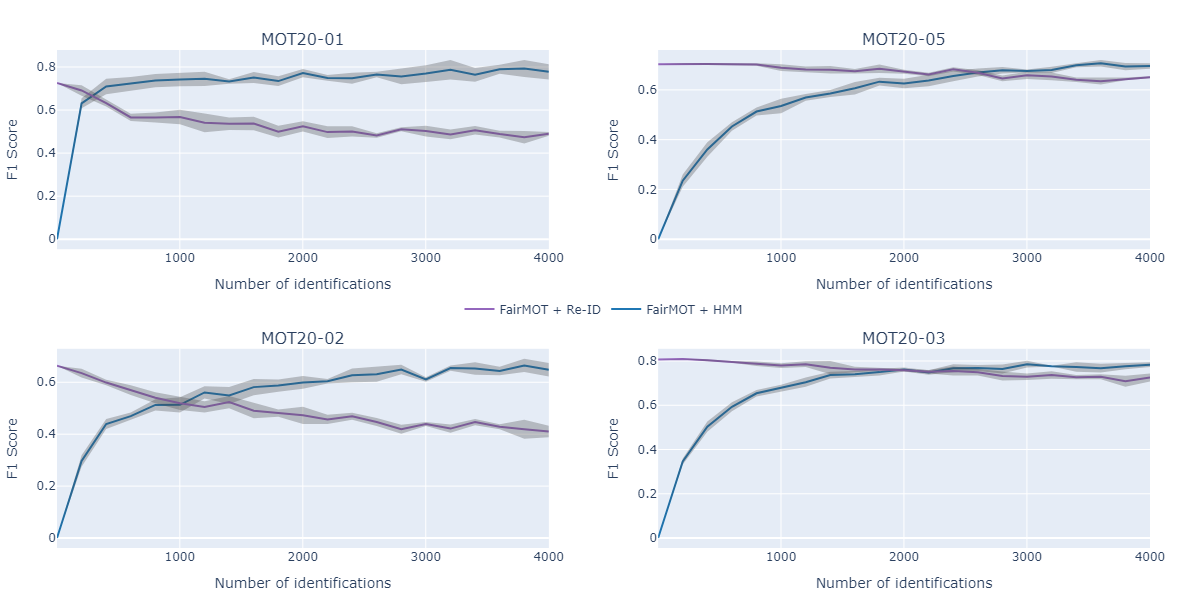}
        \caption{FairMOT as base tracker with no random observations }
        \label{fig:fairmot_no_random_mot20}
    \end{subfigure}

    \vskip\baselineskip
    \begin{subfigure}[b]{0.48\textwidth}
        \centering
        \includegraphics[width=\textwidth]{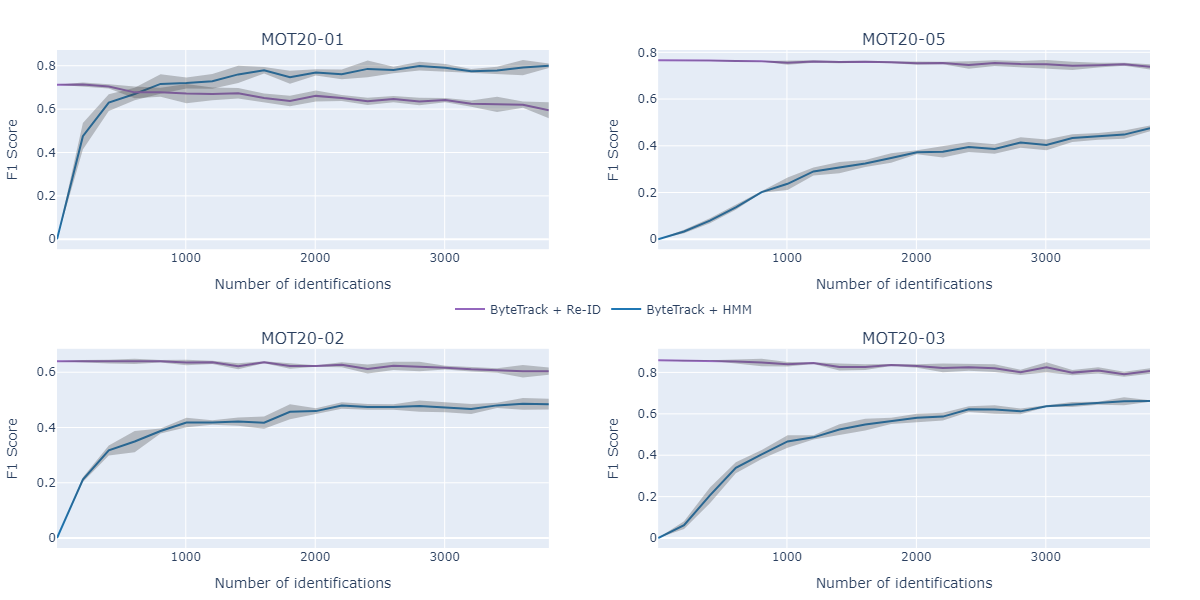}
        \caption{ByteTrack as base tracker with filtered identifications only
        }
        \label{fig:bytetrack_no_random_filtered_mot20}
    \end{subfigure}
    \hfill
    \begin{subfigure}[b]{0.48\textwidth}
        \centering
        \includegraphics[width=\textwidth]{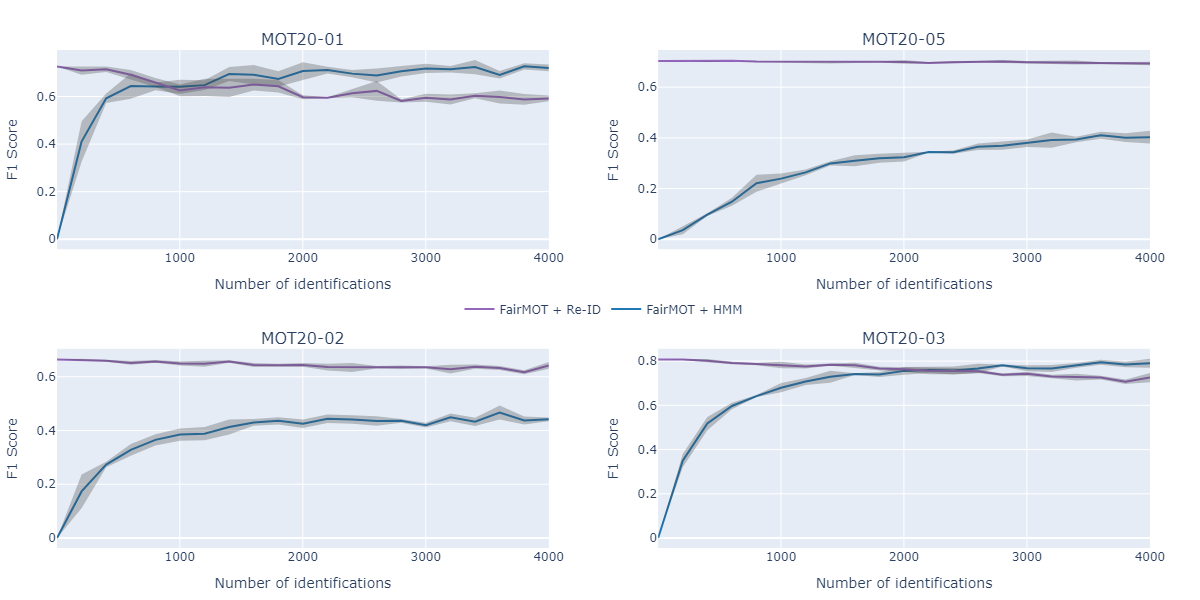}
        \caption{FairMOT as base tracker with filtered identifications only }
        \label{fig:fairmot_no_random_filtered_mot20}
    \end{subfigure}

\caption{F1 score as a function of the number of identifications from the artificial identifier in the MOT20 dataset. Error bars present standard deviation from 20 random simulations.  In \ref{fig:bytetrack_random_mot20} and \ref{fig:fairmot_random_mot20}, observations are generated such that 25\% of them are randomly distributed. These do not include identification errors caused by crowded scenes during the artificial observation generation process, so there are more than 25\% of misleading observations generated. In \ref{fig:bytetrack_no_random_mot20}  and   \ref{fig:fairmot_no_random_mot20} none of the observations are randomly distributed, but rather are all intended to be correct. In \ref{fig:bytetrack_no_random_filtered_mot20} and  \ref{fig:fairmot_no_random_filtered_mot20}, observations are filtered to remove misleading identification candidates, and no random identifications are added to limit the number of misleading observations.}

\label{fig:bytetrackmot20}
\end{figure*}

Similar observations can be made on all plots of Figures \ref{fig:bytetrackmot17} and \ref{fig:bytetrackmot20} where base trackers (ByteTrack and FairMOT) + HMM initially show lower performance but improve over the increasing number of identifications, eventually outperforming ByteTrack+Re-ID and FairMOT+Re-ID as more identifications are available, even with a significant number of misleading identifications.
 
However, compared to our 10-minute video, where the scene is less crowded and animals are less occluded, we noticed that many of the highest probabilities obtained from our artificial identifications were not assigned to the right person. Additionally, we observe a decline in the performance of FairMOT+Re-ID and ByteTrack+Re-ID when we increase the number of uncertain identifications. The consequence of this problem is the difficulty for FairMOT+Re-ID and ByteTrack+Re-ID to improve the performance of their base trackers, as shown in figures \ref{fig:bytetrack_random}, \ref{fig:fairmot_random}, \ref{fig:bytetrack_no_random}, and \ref{fig:fairmot_no_random}.
 
To reduce the effect of this problem we performed another experiment where we removed some of those misleading identifications (Figure \ref{fig:bytetrack_no_random_filtered}, Figure \ref{fig:fairmot_no_random_filtered}, Figure \ref{fig:bytetrack_no_random_filtered_mot20}, and Figure \ref{fig:fairmot_no_random_filtered_mot20}). Briefly, we added filters that require a minimal IOU of 0.7 for the detection associated to the ground truth identification and a probability higher than 0.5 for at least one detected object. This procedure does not guarantee the removal of all misleading observations, but helps mitigate at least some of them. With these filters, fewer identifications were provided to ByteTrack+HMM, FairMOT+HMM, ByteTrack+Re-ID, and FairMOT+Re-ID as we can see in figures \ref{fig:fairmot_no_random_filtered} and \ref{fig:bytetrack_no_random_filtered}. In that experiment, no random observations were introduced, and we can observe that Re-ID is able to improve the base trackers on most datasets, similarly to our HMM approach. However, we also notice that in this context, our HMM is not that effective since we are in the scenario without filtering. This is likely due to the reduced number of identifications, which confirms the importance of the number of identifications for the HMM framework.

\subsection{Discussion}

The goal of our work was to provide a solution for identity-aware long-term tracking with sparse, uncertain identity information. As shown in Table \ref{table_compar} our HMM framework improves upon the basic tracker and the classic re-identification. This means that, at a specific time $t$, by using past and future identification information, an HMM-based tracker can make better decisions on identity assignment. When examining the F1 score over time for the three tracking approaches presented in Figure \ref{fig:F1_comp}, we observe a decrease in performance over time for ByteTrack and ByteTrack+Re-ID. Meanwhile, ByteTrack+HMM remains stable and becomes more accurate over time, due to the increased number of identifications after 6K frames. We expect the performance of the HMM-based tracker to improve with the number of identifications provided by the $identifier$ (as observed in Figure \ref{fig:evol}). In our evaluation with a variable number of identifications (Figures \ref{fig:evol}, \ref{fig:bytetrackmot17} and \ref{fig:bytetrackmot20}), Re-ID-based trackers tend to stagnate with no significant improvement due to the identification errors whereas HMM-based trackers exhibit continuous improvements in the F1 score as the number of identifications increases. This is because the HMM framework makes better use of well-distributed identification information to make correct assignments despite the presence of unreliable identifications.

Figures \ref{fig:bytetrackmot17} and \ref{fig:bytetrackmot20} illustrate that the number of identifications is a crucial factor for the HMM framework, for both ByteTrack and FairMOT. The HMM framework demonstrates its ability to extract more information from a large number of uncertain identifications compared to a smaller number of more certain identifications, outperforming Re-ID in this regard. This indicates that the HMM framework is robust to identification errors, suggesting that it is a promising approach for uncertain long-term identity-aware tracking.

\section{Conclusion}

In this work, we propose a framework based on an HMM reformulation of the tracking in order to tackle the uncertain identity-aware long-term tracking problem where sparse identity information is available. This framework takes advantage of the forward-backward algorithm for information propagation to provide a solution for this problem. The HMM modeling requires two key pieces of information: probabilities linking objects to identities and a tracking matrix matching objects between consecutive frames. 
The experiments showed that the framework is better than a basic re-identification, can handle identity information error, and is more accurate if more identifications are provided, even with some mistakes, which are desirable properties for accurate long-term tracking.

While our work demonstrates that integrating sparse and uncertain external identifications can improve tracking performance, the proposed method relies on an offline approach, which limits its applicability to real-time tracking. While implementing a time window could allow for real-time use, developing a fully online version remains an important direction for future research. Although our experiments cover multiple domains, the challenges associated with sporadic identification observed in the livestock sector may similarly affect other fields. We believe this HMM framework holds promise for applications such as human video surveillance, player tracking, and robotic vision. A promising direction would be to explore integrating Re-ID models (trained to identify individuals) with typical trackers to gauge the framework’s performance in assembling such components.
\section{Acknowledgment}
This work was supported by the  MAPAQ (Quebec Ministry of Agriculture, Fisheries and Food) Innov'Action program. Thanks to the CDPQ (Centre de développement du porc du Québec) employees for gathering video data and Andréane Bélanger-Roy for annotations.

\section{Statements and Declarations}
The authors declare no competing interests.

\section{Data availability}
Data are publicly available at this link : \href{https://github.com/ngobibibnbe/uncertain-identity-aware-tracking}{https://github.com/ngobibibnbe/uncertain-identity-aware-tracking}


\end{document}